\documentclass[letterpaper, 10 pt, conference]{ieeeconf}  

\IEEEoverridecommandlockouts
\usepackage{cite}
\usepackage{amsmath,amssymb,amsfonts}
\usepackage{algorithmic}
\usepackage{graphicx}
\usepackage{textcomp}
\usepackage{xcolor}
\def\BibTeX{{\rm B\kern-.05em{\sc i\kern-.025em b}\kern-.08em
    T\kern-.1667em\lower.7ex\hbox{E}\kern-.125emX}}

\begin{document}

\title{\textbf{Fluidic FlowBots: Intelligence embodied in the characteristics of recirculating fluid flow}\\
\thanks{
M. Gepner, J. Mack, F. Giorgio-Serchi, and A.A. Stokes $($\texttt{adam.stokes@ed.ac.uk}$)$ are with The School of Engineering and The Edinburgh Centre for Robotics at The University of Edinburgh, The King's Buildings, West Mains Road, Edinburgh, Scotland, EH9 3FF.\\
The authors gratefully acknowledge the supported of the EPSRC CDT-RAS \textbf{EP/S023208/1}\\
For the purpose of open access, the author has applied a Creative Commons Attribution (CC BY) licence to any Author Accepted Manuscript version arising from this submission.
*Address all correspondence to this author.}
}

\author{{Maks Gepner, }
\and
{Jonah Mack, }
\and
{Francesco Giorgio-Serchi, \textit{Member IEEE}}
\and
{and Adam A. Stokes$^*$, \textit{Member IEEE}}
}

\maketitle

\begin{abstract}
The one-to-one mapping of control inputs to actuator outputs results in elaborate routing architectures that limit how complex fluidic soft robot behaviours can currently become. Embodied intelligence can be used as a tool to counteract this phenomenon. Control functionality can be embedded directly into actuators by leveraging the characteristics of fluid flow phenomena. Whilst prior soft robotics work has focused exclusively on actuators operating in a state of transient/no flow (constant pressure), or pulsatile/alternating flow, our work begins to explore the possibilities granted by operating in the closed-loop flow recirculation regime. Here we introduce the concept of FlowBots: soft robots that utilise the characteristics of continuous fluid flow to enable the embodiment of complex control functionality directly into the structure of the robot. FlowBots have robust, integrated, no-moving-part control systems, and these architectures enable: monolithic additive manufacturing methods, rapid prototyping, greater sustainability, and an expansive range of applications. Based on three FlowBot examples: a bidirectional actuator, a gripper, and a quadruped with a swimming gait -  we demonstrate how the characteristics of flow recirculation contribute to simplifications in fluidic analogue control architectures. We conclude by outlining our design and rapid prototyping methodology to empower others in the field to explore this new, emerging design field, and design their own FlowBots.
\end{abstract}


\section{Introduction}
\label{sec:introduction}

    \subsection{Embodied intelligence expands the capabilities of fluidic soft robots.} \label{sec:problem}
    
    The field of robotics has seen a plethora of developments in using fluid pressure (pneumatic and hydraulic) to actuate compliant, soft-bodied robots \cite{jumet_data-driven_2022}; these soft-robots exhibit superior capabilities over their rigid counterparts in use cases requiring high dexterity and safety during operation with humans \cite{polygerinos_soft_2017, whitesides_soft_2018}. Replacing electronic controls with fluidics has also opened up the possibility of deploying soft robots in extreme environments, thanks to the inherent high power density, and temperature and radiation resistance of fluidic systems \cite{mahon_soft_2019, yirmibesoglu_evaluation_2019, zhang_progress_2023}.

    There have been a range of notable advancements in recent years for controlling fluidic soft robots using hardware that mimics digital electronic devices and architectures  \cite{napp_simple_2014, wehner_integrated_2016, galloway_soft_2016, vogt_shipboard_2018, rothemund_soft_2018, preston_digital_2019, preston_soft_2019, garrad_soft_2019, wbartlett_fluidic_2020, drotman_electronics-free_2021, song_cmos-inspired_2021, gallardo_hevia_high-gain_2022, hubbard_fully_2021, lee_buckling-sheet_2022, zhai_desktop_2023, buchner_vision_2023}. Despite these advancements, scaling fluidic control system to higher degrees of complexity remains difficult; a large number of pressure control inputs is required, and routing control signals becomes an issue \cite{mahon_soft_2019, jadhav_scalable_2023}.

    Attempts at overcoming this ``routing problem'' have included fluidic demultiplexer chips \cite{wbartlett_fluidic_2020}, oscillators \cite{preston_soft_2019}, fluidic matrix circuits \cite{jadhav_scalable_2023}, as well as analogue devices that forego the need for discretisation \cite{garrad_soft_2019, decker_programmable_2022}. An alternative approach involves embodied intelligence, whereby control features can be embedded directly into the robot. By exploiting the mechanics of the interaction of the robot with its physical operating environment as a form of sensory feedback, the need for external control inputs can be reduced \cite{pfeifer_self-organization_2007, cianchetti_embodied_2021, sitti_physical_2021, mengaldo_concise_2022}.

    \subsection{Internal flow characteristics can embed functionality into the structure of the actuator.} \label{sec:quest_flow_char}
    
    In most reported-cases, and for each of the actuators cited by Section \ref{sec:problem}, the pressure is adjusted by pumping set quantities of fluid between the individual chambers of the actuator; this transition is carried out in a one-off, pulsatile, or cyclic manner. Due to the mechanical coupling between the fluid medium and the vessels/channels containing it, the behaviour of a fluidic actuator can be directly influenced by controlling the pressure of the fluid inside of it. During the pumping phase, the flow rate of the fluid is determined by its interaction with the flow channels; this mechanical coupling influences the rate of change of the pressure distribution inside the actuator, and, therefore, the deformation of the actuator. If the characteristics of the flow can be exploited, they can be used to control the actuator.

    Viscous energy losses lead to a decrease in the pressure of the flowing fluid. Previous work in the field has explored exploiting this phenomenon for control purposes \cite{matia_leveraging_2017, futran_leveraging_2018, vasios_harnessing_2020, salem_leveraging_2020, matia_dynamics_2021}. In their recent work \cite{matia_harnessing_2023}, Matia et al. used the resultant pressure asymmetries to control the gait of a walking robot. Matia et al. accurately modelled the relationship between the timing of viscous phenomena and the inertial coupling between the fluid and the actuator; therefore, they demonstrated how controlling the timing of transient fluid flows can serve as a form of embodied intelligence.
   
\begin{figure}[htbp]
        \centering
        \includegraphics[width=\linewidth]{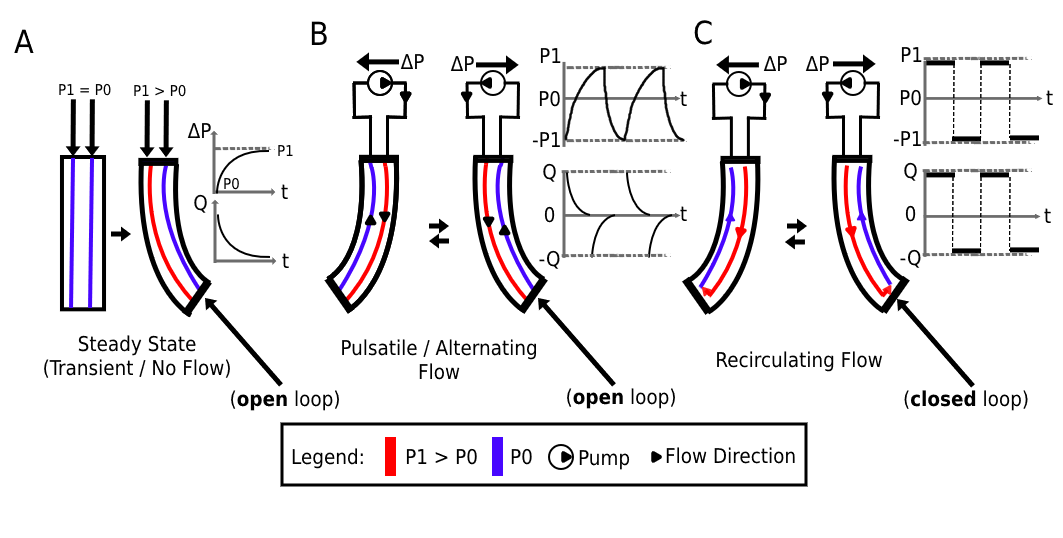}
        \caption{Soft fluidic robots utilising a) "steady state" b) "pulsatile/cyclic" c) "recirculating" flow, as defined in this paper.} 
        \label{fig:Flow_Types_Comparison}
    \end{figure}

    Whilst there as several other notable examples of previous fluidic soft robots that used ``pulsatile'', or ``cyclic'' modes of flow (such as \cite{katzschmann_cyclic_2016, wehner_integrated_2016, katzschmann_hydraulic_2016, maccurdy_printable_2016, hubbard_fully_2021, bell_modular_2022}), a condition that has not yet been explored and well-characterised in the context of soft fluidic robots is a state of continuous flow recirculation. Figure \ref{fig:Flow_Types_Comparison} illustrates the difference between soft fluidic robots based on the ``steady state'', ``pulsatile/cyclic'', and ``recirculating'' modes of flow, as defined in this paper. Fluid can be recirculated in a system for example to achieve thermal regulation, as in the work of Cacucciolo et al. \cite{cacucciolo_stretchable_2019}, and others \cite{zhang_fast-response_2019, liao_fully-printable_2021}. Liao et al. \cite{liao_fully-printable_2021} showed that the pressure of this recirculating fluid can also be used to inflate the cavities of an actuator, achieving deformation whilst simultaneously taking advantage of the thermal regulation characteristic (which they used to modify the stiffness of the actuator). An example of a strain gauge based on recirculating fluid can also be found in Koivikko et al. \cite{koivikko_integrated_2022}, although this came in the form of a separate device, and was not simultaneously used for actuation.
    
    In this paper, we delve deeper into the behaviour of soft robots operating under the closed-loop fluid recirculatory condition, describing the opportunities granted by it in the way of embodied intelligence, as well as some of the associated design challenges. We aim to empower designers to build a whole new class of fluidic robots, which we term ``FlowBots'' (as further described by Section \ref{sec:resolution_contributions}).

    \subsection{Intelligence embodied in fluid flow characteristics: from shark intestines and Tesla-valves to soft robots that can be printed as one part.} \label{sec:flueric_advantage}

    The state of ``recirculation'' is characterised by the driving fluid, taken from a reservoir, being pumped, through the channels in the actuator channels, and back to the reservoir, as illustrated on Figure \ref{fig:Flow_Types_Comparison}. To use an electrical analogy, one can think of the actuator as a source of impedance in a closed-loop fluidic network. The "current" (the driving fluid) flows from a "high potential" (outlet of the pump/pressure source), through the "impedance" (internal channels in the actuator), and back to a low potential (the fluid reservoir from which the fluid was originally pumped). Energy is lost via viscous friction, which constitutes the resistive component of the fluidic impedance, and the drop in fluid pressure that results is analogous to a drop in electrical potential.

\begin{figure}[htbp]
        \centering
        \includegraphics[width=.8\linewidth]{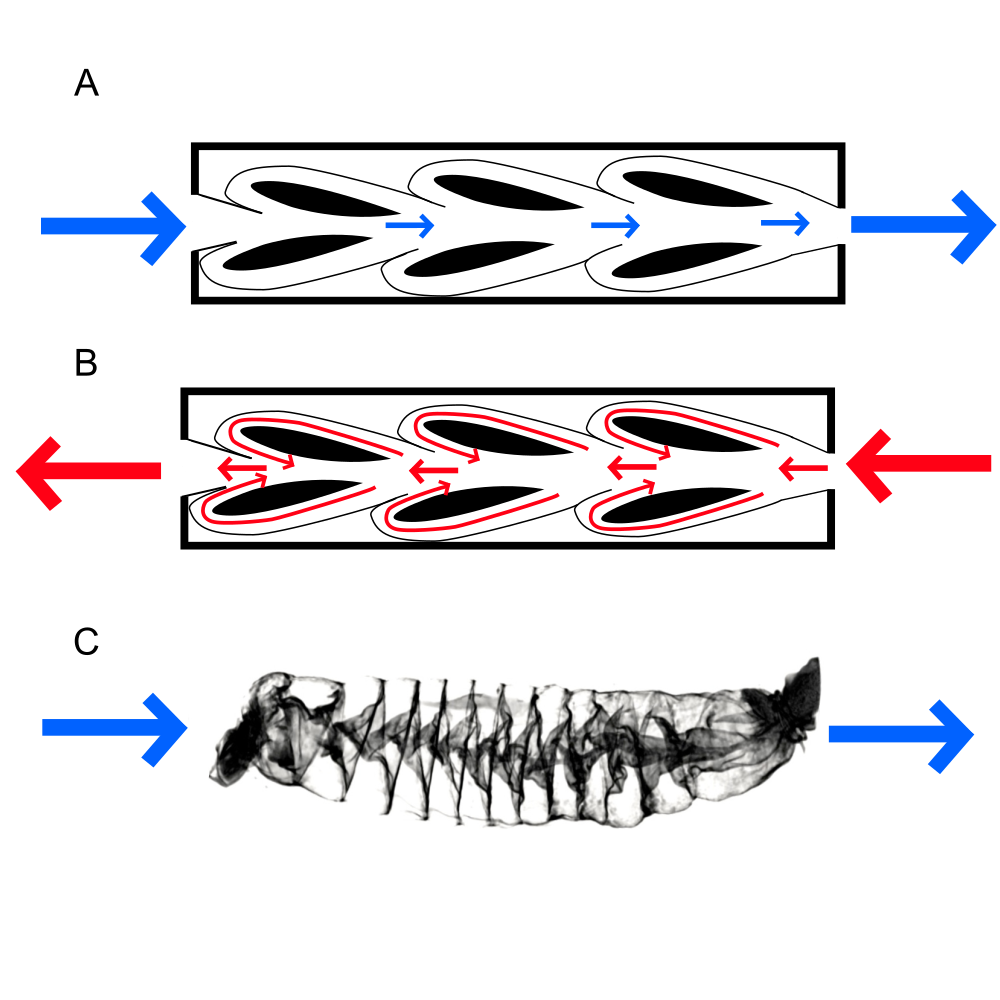}
        \caption{Inspiration for exploiting fluid flow phenomena as a form of embodied intelligence can be found in nature. The geometry of the shark intestine (c) resembles that of a Tesla-valve (a, b). Flow circulation induced by the spiral column geometry results in a diode-like behaviour, making it harder for fluid to flow in one direction (left-to-right as pictured on the figure) than the other (right-to-left). This type of fluidic diode has no moving components; the added functionality is embodied into the structure of the device. Figure (c) has been adapted from the CT scans found in \cite{leigh_shark_2021}.}
        \label{fig:Fluerics_Examples_From_Nature_And_Man_Made}
    \end{figure}

    There are also other fluid flow phenomena that can contribute to a change in the pressure distribution in the system. Some of these phenomena deal with the inertial component of the fluidic impedance, which is analogous to electrical inductance. A simple example is the Tesla-valve \cite{nikola_valvular_1920}, where the geometry of the device results in pronounced impedance in one flow direction compared to the other. As visualised by Figure \ref{fig:Fluerics_Examples_From_Nature_And_Man_Made}(a,b), the added secondary flow "loops" impede the main flow when they collide with it, as a result of the inertia they carry. In nature, a similar geometry can be found in the intestinal tracts of certain species of sharks, skates, and rays. Leigh et al. \cite{leigh_shark_2021} demonstrate that this geometry slows the flow in one direction due to the added fluidic impedance; they hypothesise that this geometry helps the animals slow the rate of food transit (thereby improving nutrient absorption in the short intestinal tract), as well as reducing the energy cost of digestion due to promoting flow in only one direction. The advantage of this type of geometry is that it avoids moving components (such as the flaps found in heart valves), which would be prone to blockage.

    \subsection{FlowBots: a new class of soft robots with intelligence embedded in the characteristics of recirculating fluid flow.} \label{sec:resolution_contributions}

    This paper focuses on embedding additional functionality directly into the bodies of soft robots, by taking advantage of the properties of recirculating fluid flow. Here, we demonstrate that new types of soft robots can be manufactured as a single part, with embedded control, and actuation. This mode of manufacture enables rapid prototyping, increases robustness, reduces waste, and minimises the risk of environmental pollution (opening up opportunities for on-site manufacturing at remote sites, and new applications in e.g. wildlife biodiversity monitoring). We term such robots ``FlowBots''. 

    FlowBots are designed to operate under the state of recirculating flow, as defined by Section \ref{sec:quest_flow_char}; this characteristic differentiates them from previous soft robots controlled and actuated using fluids. Here, fluid flows into an inlet port and flows out through an outlet port. A single actuator can have multiple inlets and outlets, as well as additional control, or sensing ports if needed. FlowBots are capable of complex behaviours with only a single pressure source, due to their embodied fluidic intelligence.

    The focus of this paper is to demonstrate the principles of operation of FlowBots based on three simple examples: a bidirectional actuator, a gripper, and a quadruped robot with a swimming gait. Each example builds upon the operating principles of the previous one. The bidirectional actuator demonstrates how pressure asymmetries arising due to viscous losses can be used for analogue control using localised flow channel constrictions. The gripper shows how flow recirculation can simplify control architectures of multi-actuator systems due to the ability of switching between connection in parallel, and in series. The quadruped FlowBot builds upon the control architecture used for the gripper, and serves as a demonstration of how the embodiment of intelligence in fluid flow characteristics allows to create robots capable of complex behaviours, yet manufactured as single, robust components, without any moving parts. Section \ref{sec:results} describes some of our experimental work in evaluating the performance of FlowBots with air and water as the operating fluid. We conclude with a discussion on some of the design challenges associated with building FlowBots, empowering others in the field to build their own.

\section{Results} \label{sec:results}

    \subsection{Localised pressure asymmetries arising due to viscous losses in fluid flow can be used for analogue, bidirectional control.} \label{sec:actuator_design}

\begin{figure}[htbp]
            \centering
            \includegraphics[width=\linewidth]{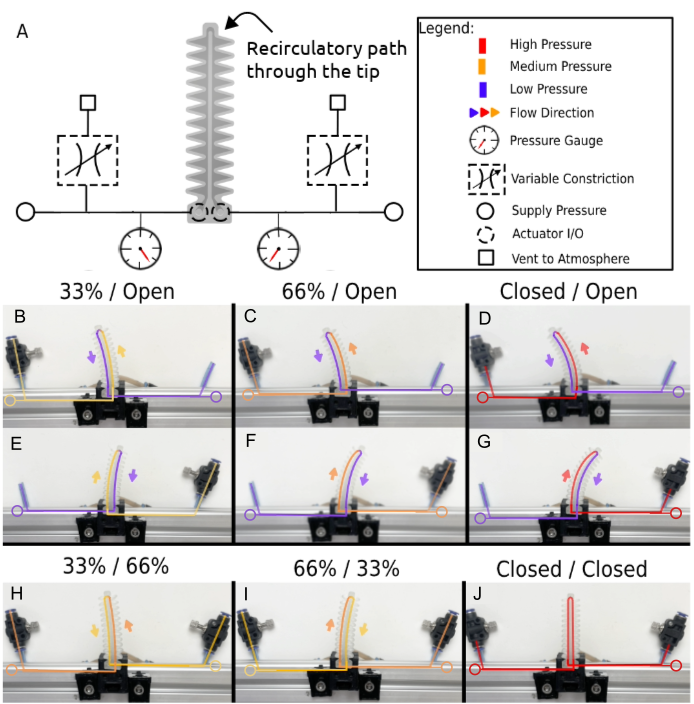}
            \caption{The principle of operation of the bidirectional actuator. By varying the outlet vent constrictions, the direction, and magnitude of the flow, as well as the pressure difference across the actuator can be controlled with an analogue characteristic. Reversing the flow direction causes the actuator to deform symmetrically in the other direction due to the mirroring of the pressure asymmetry. The percentage value of the constriction is an approximation; we estimated it indirectly by measuring the pressure difference across the actuator using two pressure gauge probes. Note that the flow path from the pressure source to the atmospheric vent is a closed, recirculatory path; this is more intuitive in the hydraulic domain, in which the reservoir, from which the flow/pressure source (PD/centrifugal pump) draws fluid, is only finitely sized.}
            \label{fig:Actuator_Operation}
        \end{figure}

        Figure \ref{fig:Actuator_Operation} illustrates the operation of the first FlowBot described in this paper -- a bidirectional actuator. The left and right pressure chambers of the actuator are connected at the tip to allow for recirculation of flow from and back to the pressure source in a closed loop. Energy is lost due to the friction between the flowing fluid and the walls of the channel, and the pressure of the fluid progressively decreases as it advances along the flow path constituted by the internal channels inside the actuator. By concentrating these viscous energy losses at the tip of the actuator, a pressure distribution is obtained that is symmetric along the  longitudinal centreline of the  actuator; the chamber in one half has high pressure, and the other half has low pressure. This pressure asymmetry results in deformation in one direction. When the flow direction is reversed, the asymmetry flips, and the actuator deforms in the opposite direction.

        If the location of the constriction was moved further upstream, or downstream of the tip, the deformation characteristic of the actuator would be permanently skewed to one side. Having the actuator deform differently in both directions at the same pressure difference could be desirable in certain applications. In this paper, all the actuators are symmetric, but we utilise external variable constrictions to regulate the direction of flow, and the pressure difference across the actuator. These constrictions can be implemented using additional valves, by manually blocking/kinking the outlet vents/tubes, or by using tactile, soft buttons (see Figure \ref{fig:tacticle_controller_object_manipulation}, as well as the Supplementary Video (available at https://wp.me/p9fxLB-Hp). Figure \ref{fig:Actuator_Operation} shows how we were able to use two such constrictions to achieve analogue control of the actuator in two directions, by regulating the pressure across it over a continuous spectrum. This analogue mapping can also be achieved by adjusting the direction of flow, as well as the flow rate of the supply pump (hydraulic case), or compressed air supply pressure (pneumatic case).

    \subsection{FlowBots have simpler control architectures due to being able to operate in parallel, as well as in series.} \label{sec:gripper}

    Section \ref{sec:actuator_design} discussed how to control a single bidirectional actuator, two controls (vent outlet constrictions) are used. To fully control two such actuators, four controls would normally be needed. By exploiting the characteristics of recirculating fluid flow, we were able to reduce this fan-in to three. The simplification comes from the fact that under the state of recirculating flow, the actuators can operate not only in a parallel, but also a series connection. The ability to switch between the two configurations allows to simplify the control architecture, and opens up the design space for constructing FlowBots consisting of several actuators, that are capable of more complex behaviours.
    
\begin{figure}[htbp]
            \centering
            \includegraphics[width=\linewidth]{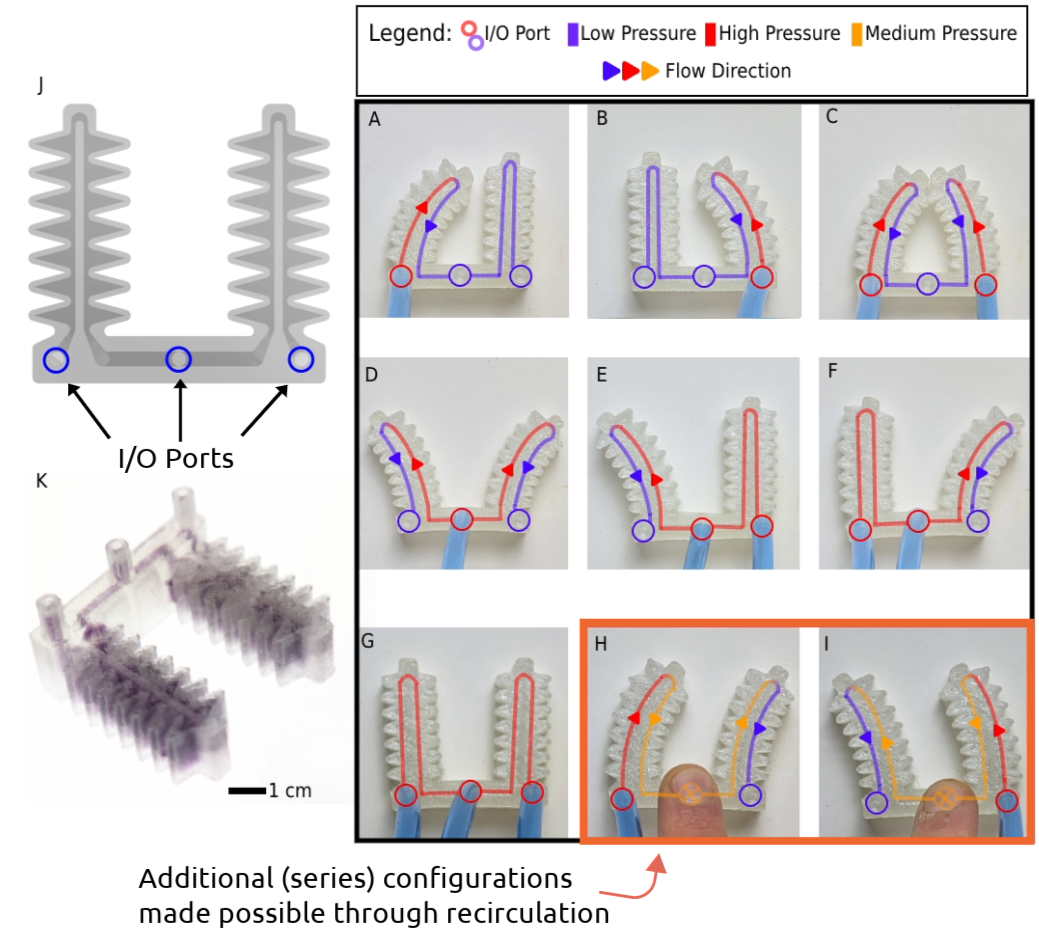}
            \caption{The principle of operation of a gripper consisting of two bidirectional actuators. By switching between a parallel ((a) to (g)) and series ((h) and (i)) connection of the actuators, we were able to achieve fully independent analogue control of each of the fingers using only three control inputs rather than four. Note that in the series configurations the actuators deform slightly less, due to the pressure difference across each one being effectively half of that in the parallel configuration.}
            \label{fig:Gripper_Operation}
        \end{figure}

    Figure \ref{fig:Gripper_Operation} illustrates the operation of the gripper that we have built using two bidirectional actuators described in Section \ref{sec:actuator_design}. The gripper uses only three control inputs rather than four, yet each of its fingers can still be controlled fully independently. The last two configurations are made possible by the fact that if the middle port is blocked, the two actuators become connected in series rather than in parallel. This capability would not be possible without using recirculating fluid flow in the system.

    \subsection{Complex FlowBots can be additively manufactured as one part, with no moving components.} \label{sec:swimmer}
\begin{figure}[htbp]
            \centering
            \includegraphics[width=\linewidth]{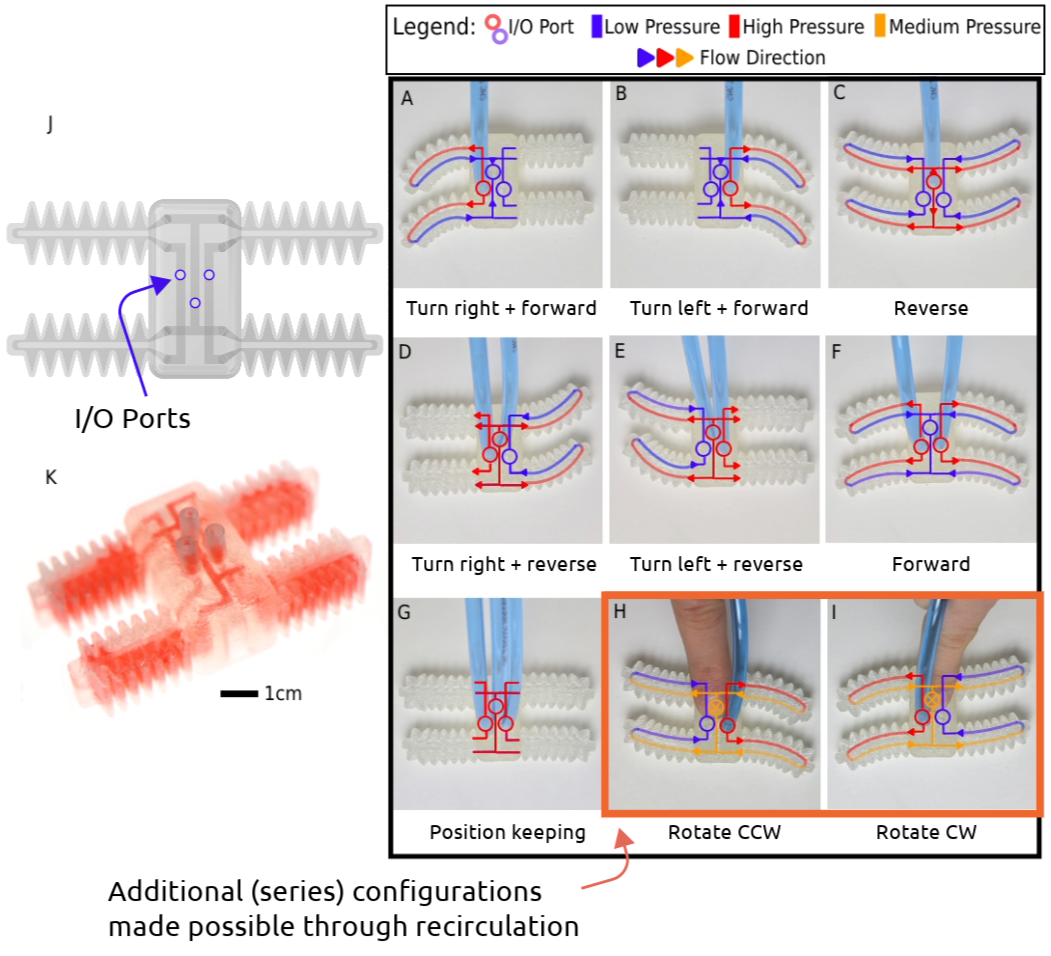}
            \caption{Operation of a quadruped FlowBot with a swimming gait that builds upon the control architecture used for the gripper described by Section \ref{sec:gripper}. Due to the embodiment of control into the body of the robot, as well as the characteristics of recirculating flow, FlowBots of increasing complexity can still be manufactured as single components.}
            \label{fig:Swimmer_Operation}
        \end{figure}

    The control characteristics of FlowBots are embodied in the interaction of the fluid flow, and the structure of the robot structure. No added parts, or moving components are required, and, as a result, entire robots can be additively manufactured as a single part, in one print job. Figure \ref{fig:Swimmer_Operation} depicts the operation of a quadruped swimming gait robot that we have built based on the control architecture used for the gripper design described in Section \ref{sec:gripper}. As in the case of the gripper, we can fully control each pair of limbs for this FlowBot using only three control inputs rather than four, thanks to the ability to switch between parallel and series connection of the actuators. Despite the added complexity of the design, the entire 3D stack of the control architecture comes packaged into the body of the robot, and has been additively manufactured as a single component. After the print job is over, the robot is ready to use. Embodying intelligence in this way makes it easy to prototype FlowBots with ever more complex 3D designs.

\begin{figure}[htbp]
            \centering
            \includegraphics[width=\linewidth]{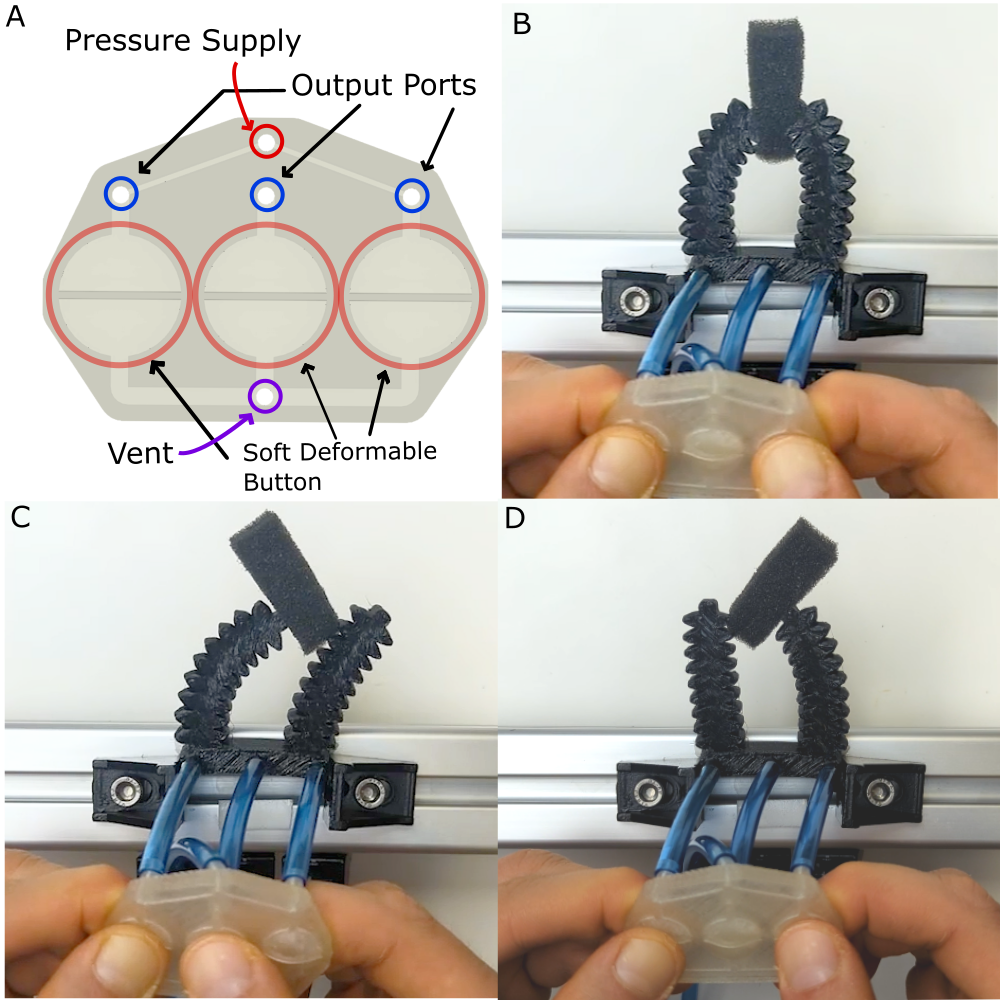}
            \caption{Manipulation of an object using the gripper design described by Section \ref{sec:gripper}. A controller with tactile, soft, buttons, enables the robot operator to control each finger of the gripper independently, by remotely adjusting the flow constrictions using the buttons. This task could also be accomplished via remotely situated (i.e. outside of the operating, potentially extreme, environment) electronic valves.}
            \label{fig:tacticle_controller_object_manipulation}
        \end{figure}

    Note that the manual connection of the tubing that routs the (same) pressure supply source to the I/O ports on on Figures \ref{fig:Actuator_Operation}, \ref{fig:Gripper_Operation}, and \ref{fig:Swimmer_Operation} is for clarity of illustration only. In Figure \ref{fig:tacticle_controller_object_manipulation}, we demonstrate how the same task can be accomplished using a tactile, analogue, soft controller that allows a remotely positioned operator to steer the FlowBot by pushing the buttons to adjust the channel restrictions. Please also refer to the Supplementary Video for a depiction of the bidirectional actuator, and swimming gait FlowBots being controlled using the same tactile controller (the former uses a 2-button variant). In an actual application scenario, tethered FlowBots could be controlled in this way, or using electronic valves in place of the buttons.

    \subsection{FlowBots can operate with water, or air, as the working fluid.} \label{sec:experiments}

\begin{figure}[htbp]
            \centering
            \includegraphics[width=\linewidth]{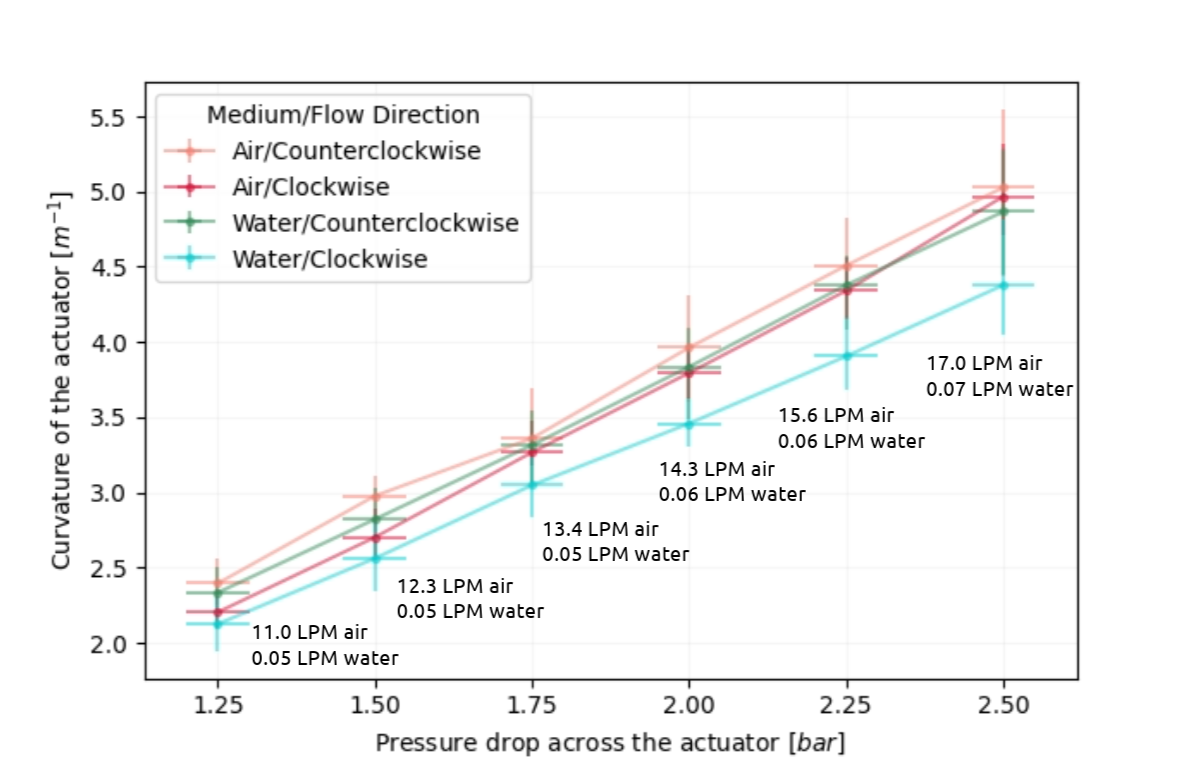}
            \caption{The bidirectional actuator deforms consistently with both incompressible (water), as well as compressible (air) media as the working fluid. We attribute the observed variations between the measurements to micro-errors in the manufacturing process on the low-cost 3D printer that we used. Note that the error bars represent the deviation in measurements between the 2 actuator specimens that were tested. The error between the 3 measurement samples gathered during each of the experimental runs was negligible. The measured fluid flow rates at each pressure are annotated on the figure, although they were not explicitly needed for this specific analysis.}
            \label{fig:Results_Deformation_And_Response_Time}
        \end{figure}

        The mechanical properties of the working fluid affect the characteristics of the recirculating flow. The fluid viscosity influences the flow rate in the system. The same flow rate with a more viscous fluid, such as water, will result in larger viscous losses, and therefore larger pressure drops than with a less viscous fluid, such as air. Nevertheless, for the same magnitude of pressure difference across an actuator, we expect the deformation to be identical, as the deformation of the bellows of the actuator is driven by the static component of the pressure. To validate this assumption, we conducted an experiment where we measured the deformation (curvature) of the bidirectional actuator described in Section \ref{sec:actuator_design} with air, and water, as the working fluid. Figure \ref{fig:Results_Deformation_And_Response_Time} shows the results of the experiment; these  results are discussed in Section \ref{sec:discussion_performance_air_water}.
        
        To account for the difference in the viscosity of the fluids, we used the pressure difference between the inlets of the actuator as the independent variable in the experiment. This approach allowed us to validate whether the actuator deforms the same amount with the same pressure difference, independent of whether air, or water, was used as the working fluid. We examined a range of pressures from 1.25 bar to 2.5 bar (in intervals of 0.25 bar), testing in both flow directions with 3 measurement repetitions per data point. We also repeated the measurements with another identical actuator specimen. Due to the symmetry resulting from placing the channel constriction at the tip of the actuator, we expect it to deform the same amount in both directions, as explained in Section \ref{sec:actuator_design}. For more details about the experimental setup, please refer to the Supplementary Material (available at https://wp.me/p9fxLB-Hp).

\section{Discussion} \label{sec:discussion}

    \subsection{FlowBots are compatible with both pneumatic, as well as hydraulic control.} \label{sec:discussion_performance_air_water}

        Figure \ref{fig:Results_Deformation_And_Response_Time} shows the analogue response of the actuator with pressure with both air and water as the working fluid. We observe that, the deformation of the actuator is consistent between the two operating media in both directions within the examined pressure range. There was a variance in the deformation between the two directions of up to 11\%; the examined specimens deform slightly more in the counterclockwise direction. Based on our experience manufacturing and testing different FlowBots, we have observed that this magnitude of error is common, and is related to micro-errors in the manufacturing process. One of the main factors at play appears to be bed levelling, which can cause the printer to extrude a different amount of material in certain locations. This phenomenon can be more pronounced on budget 3D printers such as the one used here. Due to the fact that the actuator samples used in this experiment were printed in the same location and orientation on the bed in two sequential print runs, it is possible that the bed was misaligned in one direction, or became misaligned during the print, and contributed to the error.
        
        The channels in the actuators featured in the experiment are as narrow as 0.6 mm (at the tip constriction), and variations on the order of hundreds of microns at different locations along the flow path can disturb the symmetry between the pressure distributions in both directions; such deviations can contribute to marginal differences in the deformation in two directions. In our case, these errors did not significantly affect the robot's desired functionality, but it might be a factor to consider for those designing FlowBots with very specific precision requirements. In these instances, we recommend carrying out adequate testing to ensure that the desired level of precision has been achieved.
        
        Please refer to the Supplementary Material for a discussion of our experiences with mitigating other micro-sources of variance that we have observed when carrying out our experiments, such as the hygroscopic properties of the elastic TPU filament that we used to print the robots, temperature variations in the working fluid reservoir, or small internal, or external leaks in the membranes of the actuators. 
        
        In spite of the uncertainty associated with the detected errors in measurement, our results demonstrate that FlowBots can operate consistently using both compressible, as well as incompressible media (air and water) as the driving fluid. This characteristic of cross-compatibility gives greater flexibility to designers and makes it possible to deploy FlowBots in a wider range of operating scenarios.
 
    \subsection{Recirculating flow, and facile 3D Printing enables a wide design-space and low-barrier to entry for researching and using FlowBots.} \label{sec:discussion_new_design_space}

        This paper demonstrates how designers can come up with entirely new fluidic soft robot concepts by taking advantage of the characteristics of fluid flow. Constant recirculation of the working fluid comes with the cost of energy expenditure (making the specific systems described in this paper less thermodynamically efficient than their counterparts shown in Figure \ref{fig:Flow_Types_Comparison}). but it allows to embed additional functionality in the body of the robot, simplify control architectures (as demonstrated in Section \ref{sec:gripper}), and to rapidly prototype and additively manufacture entirely new FlowBot designs. For additional design considerations to take into account when adapting systems to use recirculating flow, please refer to the Supplementary Material, where we have discussed these points in further detail. For the benefit of others, we have also open-sourced all of the CAD files for the FlowBot, as well as tactile controller, designs that have been showcased in this paper, as well as the Supplementary Video. These files can be accessed on the Soft Systems Group GitHub (https://github.com/orgs/Soft-Systems-Group); the direct link to the repository is also available on the article webpage at https://wp.me/p9fxLB-Hp.

        \textit{Recirculating flow}\\
        The exact behaviour of the fluid inside of the internal channels is difficult to characterise analytically due to the turbulence arising as a result of the combination of bellow geometries, multi-directional pipe bends, and varying channel cross-sections; this analysis is further complicated by the change in these geometries during actuator deformation, as well as the dynamic interaction of the fluid with the soft material. Whilst we do invite others specialising in the domain to help better characterise these phenomena, we encourage those interested in coming up with their own FlowBot designs to take advantage of Finite Element Method (FEM) software. Despite coming with its own difficulties arising from the same reasons as listed above, it can be a useful tool in helping identify useful trends relating design parameters, and performance. In our design methodology, we have been able to utilise Computational Fluid Dynamics (CFD) to obtain an approximation of the internal flow field that was adequate enough to facilitate effective design exploration and validation (there is also extensive work for others in refining the accuracy of these types of simulations). By combining CFD with structural simulations using Finite Element Analysis (FEA) software we were able to simultaneously explore how modifying parameters in the design affects the fluid flow in the control system, as well as the mechanical deformation of the actuators.

        \textit{3D printing}\\
        With our chosen method of Fused Deposition Modelling (FDM) 3D printing, we were able to manufacture all of the FlowBots featured in this paper as one part, without any supports or additional post-processing needed. Once taken off the print bed, each FlowBot is ready to use, as it integrates actuation, control, and even fluidic fittings in its design. In spite of the complex bellow geometry of its actuators, and the multi-layered internal 3D control architecture, the swimmer FlowBot described in Section \ref{sec:swimmer} took under 7 hours to print on a cheap, \$200 Artillery Sidewinder X1 printer using Recreus Filaflex 70A filament (as well as 82A for the specimens in Section \ref{sec:experiments}). This design approach allows for on-site manufacturing at remote sites, is more environmentally sustainable (single material monoliths are easier to dispose of), adds robustness and saves mass and space due to the reduced need for complex routing/manifolds that come with a pressurising/valving system. 

        A design variable that we have not yet explored, but which could help open up the design space even further is the observation that the fluid being recirculated throughout the system carries energy. This energy could be extracted for useful work, for example by integrating rotary tool end effectors into the robots. We foresee this could be an interesting branch to explore for those interested in developing novel FlowBot systems, in particular ones oriented toward more specific industrial applications.
      
\section{Conclusion}
\label{sec:conclusion}

    This paper introduced the concept of FlowBots, a new generation of soft robots that operate under a state of fluid flow recirculation. FlowBots are controlled by taking advantage of fluid flow characteristics, allowing for embodied intelligence. This work demonstrates the operation of three simple FlowBot systems: a bidirectional actuator, a gripper, and a quadruped robot with a swimming gait. 
    
    By utilising pressure asymmetries arising from viscous losses during fluid flow, in combination with the ability to operate in series, as well as in parallel - features only granted by flow recirculation - this work shows that we have been able to simplify the control architectures required to achieve fully independent, analogue, and bidirectional control of all of the end effectors. 
    
    Each FlowBot was additively manufactured as a monolithic part with no moving components, making it low-cost, robust, and easier to dispose of at the end of its life cycle than multi-part systems utilising different types of materials. This paper outlines an efficient design methodology that will allow engineers to rapidly prototype their own FlowBots using FEM methods, and additive manufacturing. 

    This paper lays the foundations for a new and wide design space for soft robotic systems that use recirculating flow.

\bibliographystyle{unsrt}  
\bibliography{references.bbl}

\end{document}


\title{\textbf{SUPPLEMENTARY MATERIAL\\Fluidic FlowBots: Intelligence embodied in the characteristics of recirculating fluid flow}\\
\thanks{
M. Gepner, J. Mack, F. Giorgio-Serchi, and A.A. Stokes $($\texttt{adam.stokes@ed.ac.uk}$)$ are with The School of Engineering and The Edinburgh Centre for Robotics at The University of Edinburgh, The King's Buildings, West Mains Road, Edinburgh, Scotland, EH9 3FF.\\
The authors gratefully acknowledge the supported of the EPSRC CDT-RAS \textbf{EP/S023208/1}\\
*Address all correspondence to this author.}
}

\author{{Maks Gepner, }
\and
{Jonah Mack, }
\and
{Francesco Giorgio-Serchi, \textit{Member IEEE}}
\and
{and Adam A. Stokes$^*$, \textit{Member IEEE}}
}

\maketitle

\section{Design considerations for adapting systems to use recirculating flow.}
\label{appendix1}

\subsection{Deviations in flowpath consistency can alter internal pressure distributions.} \label{appendix1:flowpath_deviations}

\begin{figure}[htbp]
    \centering
    \includegraphics[width=\linewidth]{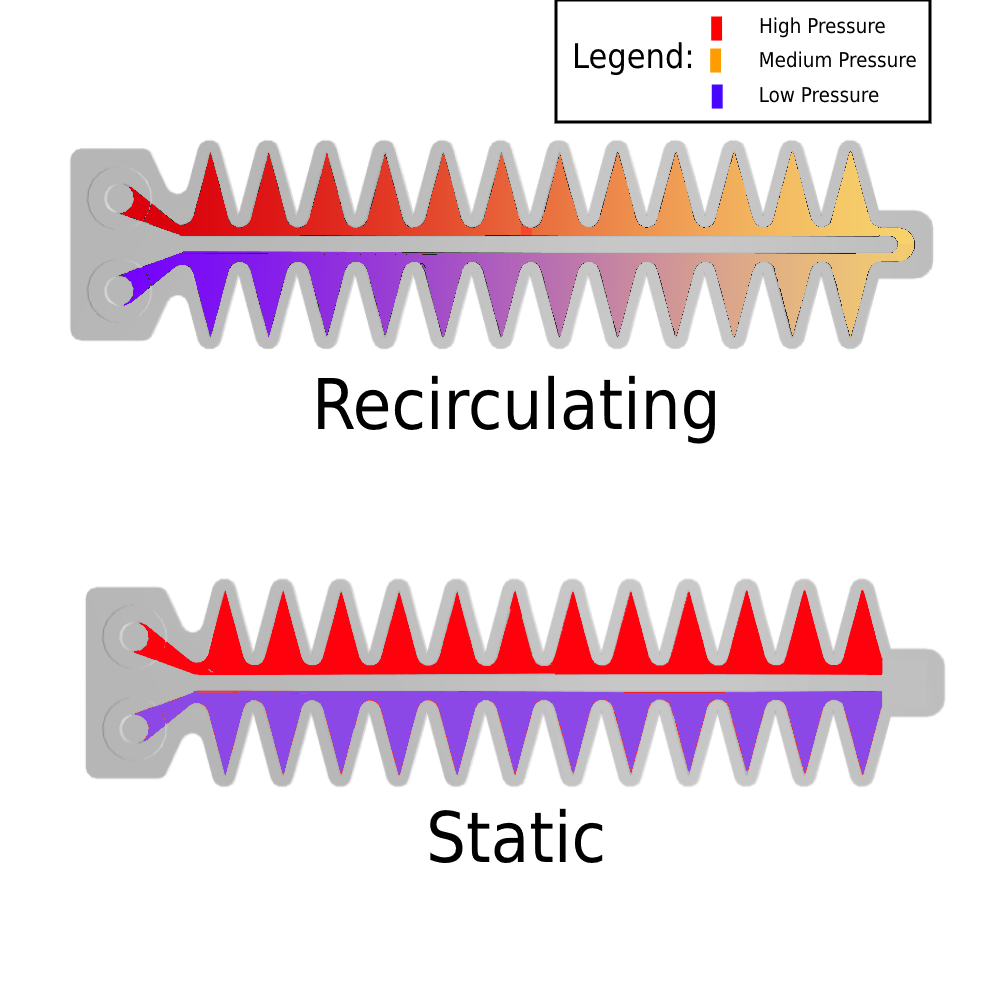}
    \caption{Imperfections in the flowpath geometry can result in slight deviations in the pressure distributions of the recirculating actuator variant (a) from the static one (b). Sources of these imperfections include manufacturing error, but also the "parasitic" resistance of the flow path. The uncertainty introduced by these factors is something to consider when designing systems that use recirculating fluid flow. Note that the colour gradient on this figure is exaggerated, to illustrate this "imperfect" case.} 
    \label{fig:Comparison_Actuator_Variants}
\end{figure}

Figure \ref{fig:Comparison_Actuator_Variants} shows two variants of the bidirectional actuator featured in the main document: one with a closed-loop path to enable recirculation of fluid, and the other being a more conventional "non-recirculating" (later described as "static") case, where the pressure chambers on the two sides are disjointed. The two chambers of the recirculating actuator variant (Figure \ref{fig:Comparison_Actuator_Variants}a) are connected through a narrow channel constriction. Ideally, all of the viscous energy losses are concentrated at this point; the pressure distribution is then identical to the static variant of the actuator (Figure \ref{fig:Comparison_Actuator_Variants}b). In reality, pressure losses at other points along the flowpath cause slight deviations in the pressure distribution. This effect can be exacerbated by micro-errors in the manufacturing process, which can cause non-uniforms widths of the channel at different points along the flowpath, or, in extreme cases, pieces of debris causing partial blockage of the channels. The latter usually results in very noticeable changes in actuator behaviour. In some cases, for example, a piece of debris blocks the channel only in one flow direction, but not the other, contributing to an asymmetry in the deformation of the actuator (the actuator deforms more in one direction than the other). Micro-leaks are also possible, although the actuators that we have examined appeared to be almost entirely leaktight beyond the pressure range examined (the maximum we tried was 4 bar).

An occasional minor point of leakage was sometimes around the port area, resulting from an imperfect fit with the tubing attached to the actuator. Note that the 4.5 mm OD ports were toleranced for a friction fit with the 4 mm ID PU tubing we used; this worked great for us, with virtually no leakage up to the working pressure of 2.5 bar, and even up to 4 bar in many cases. We were also able to print hose barb designs directly onto the parts, but the cylindrical geometry is easier to print, and connect/disconnect, so that is our recommended option. Leaks around the ports can slightly disrupt the accuracy the pressure gradient measurements between the ports, since they are located downstream of the pressure gauges. We recommend to test the tolerancing of the ports well before commencing any experimentation, to minimise the uncertainty associated with any result deviations.

Note that in recirculating variants of actuators, internal leaks can be as significant as external ones. In the case of the bidirectional actuator, any leaks through the middle, dividing membrane can also disrupt the pressure distribution.

Other factors that affect the deformation of the actuator are discussed in Appendix \ref{appendix2:other_error_sources}, along with the steps we took to control their influence. These factors include the fluid reservoir temperature (which can become altered as a result of the need to constantly actuate the fluid pressure source such as a liquid pump), as well as features of the flowpath that are downstream, and upstream of the actuator. The vector of localised fluid flow direction is a new design parameter that can be exploited to build Flowbots with new behaviours, as well as control characteristics, but it is also a source of uncertainty that should be accounted for to achieve predictable results. One example of such uncertainty is the angle at which tubes are attached to the inlet ports of the actuators. This angle can slightly influence the fluid flow interaction inside of the actuator, and alter the pressure distribution. The same can happen when the inlet port becomes slightly kinked, which changes the channel constriction and introduces an additional pressure drop at that location. Care should also be taken to ensure that the cylindrical port doesn't buckle when tubing is inserted into it, as that would narrow the channel constriction, and cause additional pressure drop that is not accounted for by the pressure sensor locations.

\subsection{Consistency in performance across the recirculating, and static actuator variants.} \label{appendix1:experimental_results}

In addition to the experimental data described in the main paper, we simultaneously gathered data on the "static" variant of the same actuator, i.e. one in which the tip constriction is sealed, to completely prevent recirculation between the two chambers (as illustrated earlier in Figure \ref{fig:Comparison_Actuator_Variants}b). We were then able to evaluate whether switching from a static to recirculating configuration affects the actuator in a measurable way. In the subsections below, we discuss the effects of this design change on the deformation of the actuator, as well as the speed of its response. For a description of the experimental setup, as well as the data processing methodology, please refer to Appendices \ref{appendix2:experimental_setup} and \ref{appendix2:data_extraction}.

\subsubsection{Deformation of the actuator} \label{appendix1:curvature_results}
\begin{figure}[htbp]
    \centering
    \includegraphics[width=\linewidth]{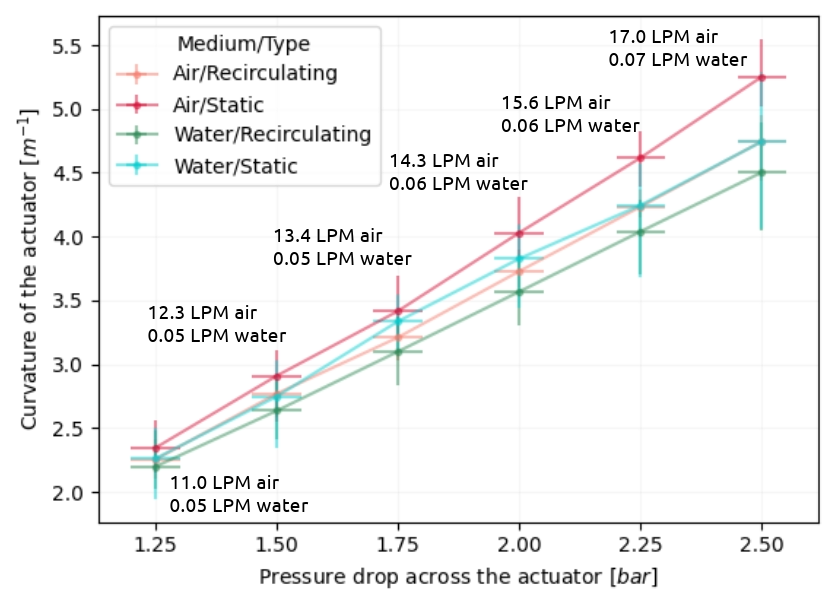}
    \caption{Our results indicate that the deformation of the recirculating actuator is consistent with both working fluids, although it is marginally smaller than the static variant; the maximum observed deviation is 13\% (with water, at 1.5 bar). As described in Appendix \ref{appendix1:flowpath_deviations}, this difference is a result of micro-deviations in the manufacturing process that alter the pressure distribution along the flowpath from the idealised case.} 
    \label{fig:Results_Curvature_TypeGroups}
\end{figure}

Figure \ref{fig:Results_Curvature_TypeGroups} shows the results of the comparison of the recirculating, and static actuator variants in terms of the deformation at different pressures. We define the deformation in terms of curvature of a circular arc created by the deformed actuator. Our results show that with both incompressible (water), as well as compressible (air) fluidic media as the working fluid, the performance of the two actuators is very consistent, although the deformation of the recirculating variant is marginally lower. This difference was below 9\% (reaching a maximum of 13\% in the case of only one datapoint). As described in Appendix \ref{appendix1:flowpath_deviations}, the error is a result of parasitic channel resistance, as well as micro-errors in the manufacturing process, which cause the internal pressure distribution of the recirculating variant to deviate from the idealised case (which would be identical to the static variant). This deviation translates into a slight change in the maximum curvature achieved by the actuator. There are also added recirculation patterns that are induced in the bellowed geometries, as well as at any sharp bends in the path of the fluid, such as at the tip of the actuator. In theory, these secondary flow patterns can also cause further deviations in the pressure distribution in the actuator. The results of our CFD analysis, however, indicated that these patterns would only have a negligible affect (relative to the large resistance caused by the channel restriction), that would not cause deviations on the order of the the ones we observed, even when the bellows are inflated during actuator deformation. The study of the flow patterns in dynamically deforming FlowBot actuators is, however, a fascinating topic for further study.

\subsubsection{Response time of the actuator} \label{appendix1:response_time_results}

\begin{figure}[htbp]
    \centering
    \includegraphics[width=\linewidth]{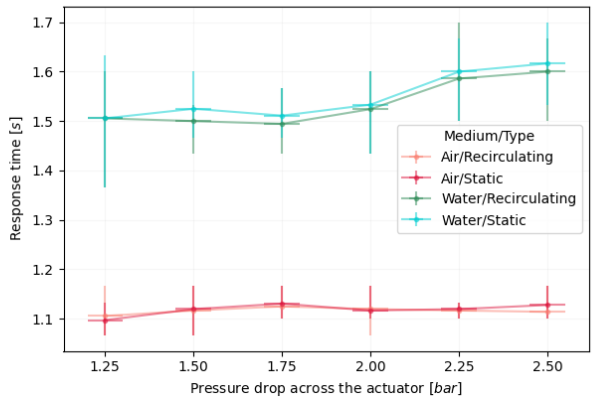}
    \caption{Adjusting the actuator to use flow recirculation does not alter the speed of its response. The response time is consistent (maximum of around 3\% error, again with water at 1.5 bar) with both air, and water as the working fluid.} 
    \label{fig:Results_Response_TypeGroups}
\end{figure}

Figure \ref{fig:Results_Response_TypeGroups} shows the comparison between the two variants of the actuator in terms of the response time (time from the start of the deformation to the moment the maximum deformation is achieved -- see Appendix \ref{appendix2:data_extraction} for more details). The experimental results indicate that the response time of the actuator was not affected in the same way as the curvature, with the maximum error being only around 3\%. This is understandable, given the trend indicated by the two lines in the figure. The response time increases only very slightly across the pressure range that was investigated (under 7\% for water, and around 2\% for air). As the actuator starts to deform, the fluid fills the additional internal volume created after the walls of the actuator are deformed. It takes longer for the denser, more viscous liquid water to accomplish this than for the compressible air medium, resulting in the response time being around 37\% longer with water than with air. Nevertheless, the additional volume is not that much larger at 2.5 bar than it is at 1.25 bar; therefore, the change in response time within this pressure range is only marginal. As a result, any deviations in curvature between the actuator variants do not translate into a significant deviation in response time from the "baseline" response time of deformation related to the delay caused by the speed of travel of the fluid in the channels.

\section{Characterising the behaviour of the actuators using high framerate video motion capture.}
\label{appendix2}

\subsection{Description of the experimental setup.} \label{appendix2:experimental_setup}

\begin{figure}[htbp]
    \centering
    \includegraphics[width=\linewidth]{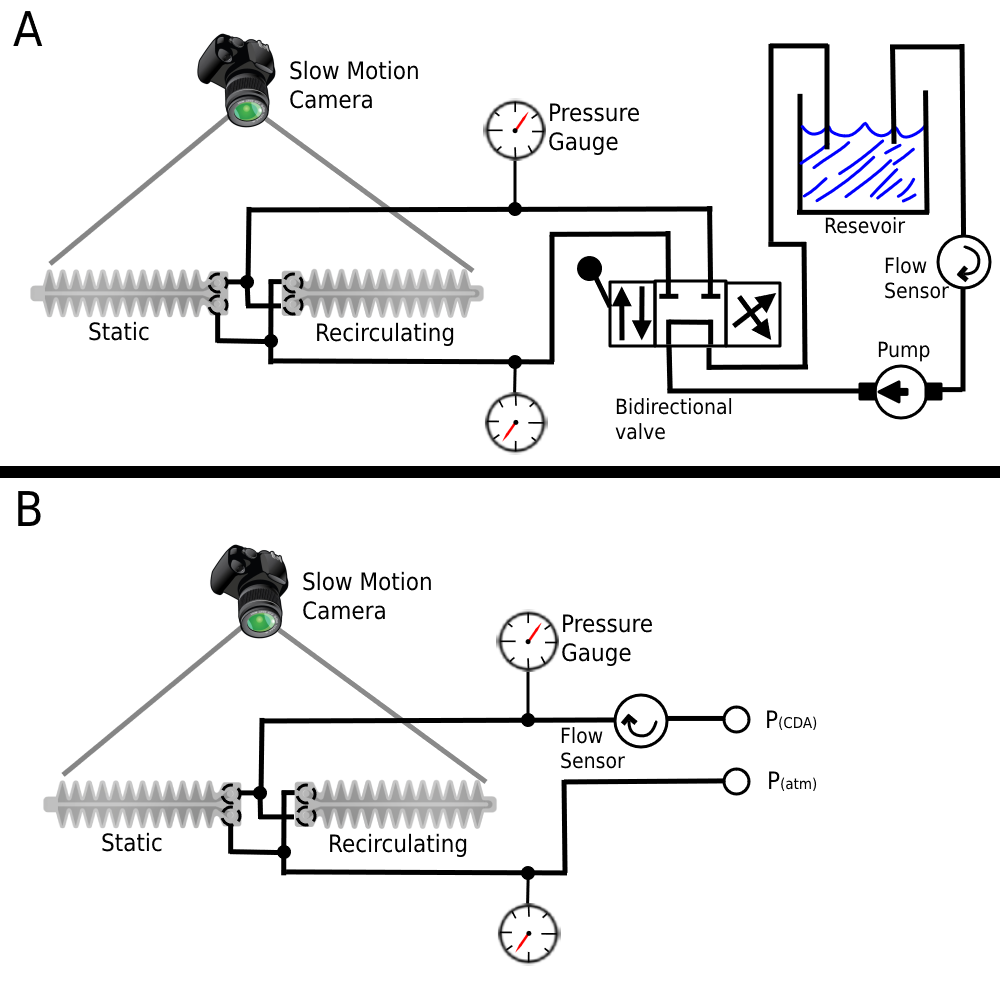}
    \caption{Experimental setup used to evaluate the deformation, and time responses of the recirculating and static actuators. In the hydraulic case (a), we controlled the PD pump flow rate to control the pressure between the actuator inlets. In the pneumatic case (b), we controlled the pressure using a CDA supply that resulted in a specified flow rate.} 
    \label{fig:Test_Setups}
\end{figure}

Figure \ref{fig:Test_Setups} shows a diagram of the experimental setup used to collect the data described in the main document, as well as Appendices \ref{appendix1:curvature_results} and \ref{appendix1:response_time_results}. To account for the effect of gravity, the two actuators were placed in a horizontally level plane and held in place using a vise. Their inlet ports were connected to each other, to ensure that the inlet pressures were consistent, as well as to isolate for minor deviations due to other factors described in Appendix \ref{appendix2:other_error_sources}. This parallel connection also ensured that the actuators deformed in the same direction relative to their own frame of reference.

In the hydraulic test setup, we used a TCS Micropumps MGD3000-F positive displacement (PD), self-priming pump to recirculate the fluid in a closed loop: from the water reservoir, through the recirculating actuator, and back to the reservoir. A Brevini lever-operated 4/2 flow directional hydraulic valve was used as the "on" switch. The PD pump outputs a set flow rate; it has to work against the fluidic resistance of the flowpath, which results in a specified value of pressure in the system. We adjust the power delivered to the pump to control its flow rate, and thus regulate the pressure across the inlets of both of the actuators.

In the pneumatic setup, a well-maintained compressed dry air (CDA) supply is used as the pressure source. A pressure regulator is used to adjust the pressure in the system. The fluidic resistance of the flowpath now instead determines the flow rate of the fluid.

To standardise between the two experimental setups, we defined the independent variable to be the pressure between the actuator inlets. We used two sensors to take probing measurements at the inlet and outlet, and then calculated the difference between the values. For our own records, we also used flow sensors (RS Pro Oval Gear hydraulic flow meter and a Sensirion SFM4300 air flow meter) to measure the fluid flow rate in the network. The flow rate measurements are annotated on Figures \ref{fig:Results_Curvature_TypeGroups} and \ref{fig:Results_Response_TypeGroups}, as well as Figure 6 of the main document, but they were not needed in our analysis.

\begin{figure}[htbp]
    \centering
    \includegraphics[width=\linewidth]{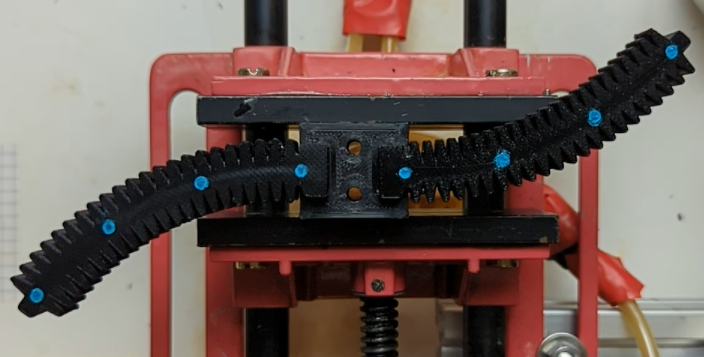}
    \caption{Example of a frame extracted from one of the videos recorded using the high-framerate sensor. The tracking dots used to estimate the actuator curvature are clearly visible in blue.} 
    \label{fig:Video_Frame_Example}
\end{figure}

As the pressure was applied, we used the 240 frames-per-second slow motion sensor of a Pixel 6 phone to capture the motion of the actuators. Each of the actuators was marked with 4 tracking dots, as visible on Figure \ref{fig:Video_Frame_Example}, which shows an example of a single frame from one of the videos we took. By tracking the position of the dots over time, we were able to estimate not only the actuator curvature, but also the speed of its response. The processing methodology that we used to extract this data is discussed in more detail in Appendix \ref{appendix2:data_extraction} below. Note that we kept the lighting consistent throughout the data gathering process, to minimise any interference with the image processing; to accomplish this, we used an overhead, diffused spotlight.

\subsection{Method of extracting data from video frames.} \label{appendix2:data_extraction}

Our processing script iterated through each frame of the slow-motion video. For each frame, we apply a colour mask to filter anything everything except for the colour of the tracking dots. This colour (blue) was selected to enable easy filtering from the background (since some fragments of the fluidic fittings we used were blue, we covered them with red tape). We record the horizontal and vertical positions of the centroids of each tracking dot, at that particular time point. The pixel value of this position is converted into millimeters based on a reference length included in the background (5mm grid paper). To help filter out some of the noise, we converted the colour from rgb to hsv, and also applied a slight gaussian blur to smooth out the tracking dot contour tracking.

\begin{figure}[htbp]
    \centering
    \includegraphics[width=\linewidth]{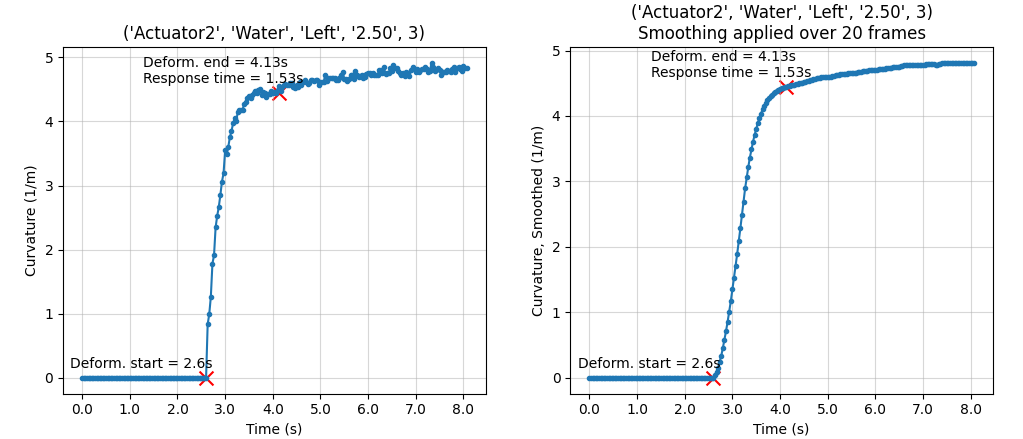}
    \caption{Example of curvature time series data extracted from the recording of the deformation of the recirculating actuator (specimen 2, with flow in the counterclockwise direction, pressure drop 2.5 bar, data sample 3). The raw data was smoothed by interpolating over a window of 20 video frames to filter out low-level noise related to the resolution of the sensor, as well as the colour masking operation on each of the video frames.} 
    \label{fig:Detailed_Response_Example}
\end{figure}

We used SciPy's .minimize method to fit the equation of a circular arc to the 4 points. In this way, we obtained an estimate of the radius of curvature of the actuator over time. Figure \ref{fig:Detailed_Response_Example} shows an example of one of these time series. To each series, we filtered out some of the low-level noise using SciPy's .interpolate method, applied over a smoothing window of 20 frames.

Figure \ref{fig:Detailed_Response_Example} also shows that the deformation characteristic of the actuators consists of an initial sharp spike followed by slow flattening out of the curve, which we attribute to the viscoelastic properties of elastomer used to manufacture the actuators (Recreus Filaflex 82A TPU filament). To ensure we set the "end" point of the deformation evenly throughout the samples, we defined the cutoff threshold as the time when the rate of deformation of the actuator drops below 5\% over a 0.4s rolling time window. The difference between the deformation start and end times was then the "response time". The extracted curvature values were taken as the curvature at that "end" time point.

To reduce measurement error, we repeated each measurement 3 times (restarting the pressure source and setting it to the same value), and we repeated the experiment with an additional specimen of each of the actuators to account for the influence of manufacturing error. Although it would have been possible to reduce this error further by examining even more specimens, we decided to include just these two in the final data, to highlight the importance of this factor in FlowBot design (as mentioned in the main document, we have noticed that the magnitude of error that was observed was common across other specimens that we have examined). As outlined in Appendix \ref{appendix1:flowpath_deviations}, manufacturing defects can influence the behaviour of FlowBots, and are something that designers looking for maximum precision should account for. At the same time, varying internal geometries are also a parameter that can be exploited to come up with new FlowBot designs. For example, the piece of debris described in Appendix \ref{appendix1:flowpath_deviations} could be deposited deliberately, to cause asymmetric deformation behaviour of the actuator, perhaps within a specified flow rate/pressure range.

\subsection{Mitigation of other sources of experimental error.} \label{appendix2:other_error_sources}
To mitigate the error resulting from the variables described in Appendix \ref{appendix1:flowpath_deviations}, we took the following steps:

\subsubsection{Fluid/actuator temperature}
The thermal capacity of the fluid reservoirs was kept very large, to minimise temperature deviations that can affect the mechanical properties of the extremely soft elastomer material from which the actuator is made. In extreme cases, the temperature could also affect the properties of the working fluid. We used a 10L reservoir for water, and a well-maintained Compressed Dry Air (CDA) supply in the pneumatic case, which we deemed to have an adequately large capacity (we also vented the air directly to the atmosphere). We also monitored the temperature of the fluid using a DS18B20 temperature probe, to ensure consistency throughout the whole experiment.

\subsubsection{Manufacturing errors}
The actuators described in the experiment were manufactured on the same 3D printer (Artillery Sidewinder X1), and using the same material (Recreus Filaflex 82A), as well as print settings (0.2mm layer height, 40mm/s speed). The filament was also dried in a heated chamber before printing to minimise the effect of any residual moisture that could affect the printing consistency. The recirculating, and static variants were printed as a single, sequential print job, resulting in them being placed on slightly different locations on the print bed, although we printed the 2 specimens of each actuator type in exactly the same location. Imperfections in the levelling of the print bed can result in micro-scale printing errors; if a higher degree of consistency between specimens is required, we recommend to pay additional attention to bed levelling in between prints. To mitigate the impact of small leaks, we pressurised each of the actuator specimens up to 4 bar whilst underwater, and visually observed the location of any air bubbles. Samples with excessive leakage were re-printed. For the specimens discussed in the paper and this supplementary material, leaks did not appear to be a problem.

\subsubsection{Fluid insertion angle}
To make sure the angle at which the fluid enters the actuator is consistent, we printed a special jig that holds the inlet tubes in place (making sure they are as straight as possible). The fluid pressure probes were also kept as close as possible to the actuator inlets, to make sure that the reading is representative of the pressure at those points (the parasitic fluidic resistance of the tubes, as well as any twists, turns, leaks, and constrictions, could cause slight deviations that would not be detectable using the pressure probes).
